\documentclass{article} 
\usepackage{iclr2021_conference,times}

\usepackage{hyperref}
\usepackage{url}
\usepackage{xcolor}
\usepackage{multirow}
\usepackage{amsmath}
\usepackage[ruled,vlined]{algorithm2e}
\usepackage{graphicx}
\usepackage{xspace}
\usepackage{wrapfig}
\usepackage[normalem]{ulem}
\usepackage{tablefootnote}

\title{Scalable Graph Neural Networks for \\Heterogeneous Graphs}

\author{Lingfan Yu\thanks{Work done while Lingfan Yu was an intern at Facebook AI Research.} \\
New York University\\
\texttt{lingfan.yu@nyu.edu}
\\
\And
Jiajun Shen \\
Facebook AI Research \\
\texttt{jiajunshen@fb.com} \\
\And
Jinyang Li \\
New York University\\
\texttt{jinyang@cs.nyu.edu } \\
\And
Adam Lerer \\
Facebook AI Research \\
\texttt{alerer@fb.com}
}

\def\hn{\sffamily\selectfont}
\newcommand{\mpfont}{\hn\scriptsize}
\ifx\nocolormarks\undefined
\newcommand{\MPworker}[2]{\unskip{\color{#1}\vrule\vrule\vrule\vrule}{\marginpar{\vspace{-2em}\raggedright\color{#1}\mpfont #2}}}
\else
\newcommand{\MPworker}[2]{\unskip}
\fi

\ifx\nocolormarks\undefined
  
\else
  
\fi

\newcommand{\cellalign}[2]{\multicolumn{1}{#1}{#2}}

\newcommand{\approach}{NARS\xspace}
\newcommand{\approachfull}{Neighbor Averaging over Relation Subgraphs\xspace}

\iclrfinalcopy 
\begin{document}

\maketitle


\begin{abstract}
Graph neural networks (GNNs) are a popular class of parametric model for
    learning over graph-structured data. Recent work has argued that GNNs
    primarily use the graph for feature smoothing, and have shown
    competitive results on benchmark tasks by simply operating on graph-smoothed
    node features, rather than using end-to-end learned feature hierarchies that
    are challenging to scale to large graphs. In this work, we ask whether these
    results can be extended to \textit{heterogeneous} graphs, which encode
    multiple types of relationship between different entities. We propose
    \textit{\approachfull} (\approach), which trains a classifier on
    neighbor-averaged features for randomly-sampled subgraphs of the ``metagraph''
    of relations. We describe optimizations to allow these sets of node features
    to be computed in a memory-efficient way, both at training and inference
    time. \approach achieves a new state of the art accuracy on several benchmark
    datasets, outperforming more expensive GNN-based
    methods.
\end{abstract}

\section{Introduction}

In recent years, deep learning on graphs has attracted a great deal of interest, with new applications ranging from social networks and recommender systems, to biomedicine, scene understanding, and modeling of physics \citep{wu2020comprehensive}. One popular branch of graph learning is based on the idea of stacking learned ``graph convolutional" layers that perform feature transformation and neighbor aggregation \citep{kipf2016semi}, and has led to an explosion of variants collectively referred to as Graph Neural Networks (GNNs) \citep{GraphSAGE,xu2018GIN,Velickovic2018GAT}. Most benchmarks for learning on graphs focus on very small graphs, but the relevance of such models to large-scale social network and e-commerce datasets was quickly recognized \citep{pinsage}. Since the computational cost of training and inference on GNNs scales poorly to large graphs, a number of sampling approaches have been proposed that improve the time and memory cost of GNNs by operating on subsets of graph nodes or edges \citep{GraphSAGE,chen2017stochastic,zou2019layersampling,zeng2019graphsaint,chiang2019cluster}.

Recently several papers have argued that on a range of benchmark tasks -- social network and e-commerce tasks in particular -- GNNs primarily derive their benefits from performing feature smoothing over graph neighborhoods rather than learning non-linear hierarchies of features as implied by the analogy to CNNs \citep{wu2019SGC,nt2019gfNN,chen2019powerful,rossi2020sign}. Surprisingly, \cite{rossi2020sign} demonstrate that a one-layer MLP operating on concatenated N-hop averaged features, which they call Scalable Inception Graph Network (SIGN), performs competitively with state-of-the-art GNNs on large web datasets
while being more scalable and simpler to use than sampling approaches. Neighbor-averaged features can be precomputed, reducing GNN training and inference to a standard classification task.

However, in practice the large graphs used in web-scale classification problems are often heterogeneous, encoding many types of relationship between different entities \citep{lerer2019pytorch}. While GNNs extend naturally to these multi-relation graphs \citep{schlichtkrull2018modeling} and specialized methods further improve the state-of-the-art on them \citep{hu2020heterogeneous,wang2019heterogeneous}, it is not clear how to extend neighbor-averaging approaches like SIGN to these graphs.

In this work, we investigate whether neighbor-averaging approaches can be
applied to heterogeneous graphs (HGs). We propose \textit{\approachfull} (\approach), which computes neighbor averaged features for random subsets of relation types, and combines them into a single set of features for a classifier using a 1D convolution. We find that this scalable approach exceeds the accuracy of state-of-the-art GNN methods for heterogeneous graphs on tasks in three benchmark datasets.

One downside of \approach is that it requires a large amount of memory to store node features for many random subgraphs. We describe an approximate version that fixes the memory scaling issue, and show that it does not degrade accuracy on benchmark tasks.

\begin{figure}
    \centering
    \includegraphics[width=\textwidth]{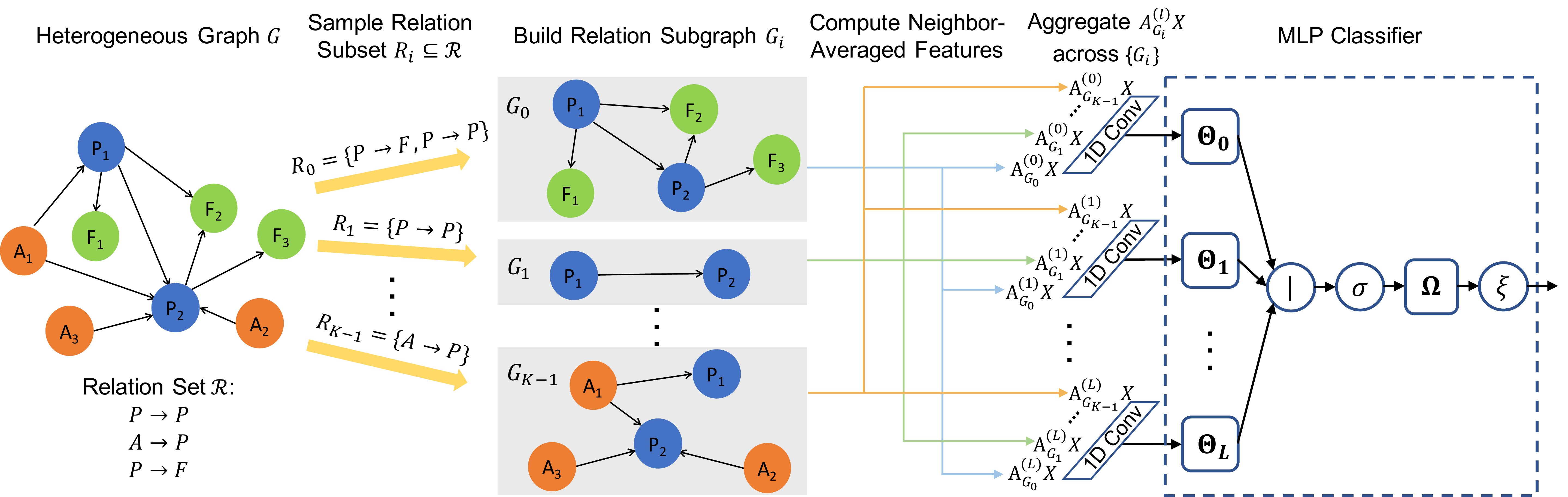}
    \caption{\small \approachfull on heterogeneous graph $G$. $G$ has three node
     types: Paper (P), Author (A), and Field (F), and three relation types:
     Paper cites Paper (P$\rightarrow$ P), Paper belongs-to Field
     (P$\rightarrow$F), Author writes Paper (A$\rightarrow$P).}
    \label{fig:overview}
\end{figure}

\section{Background}

Graph Neural Networks are a type of neural model for graph data that uses graph structure to transform input node features into a more predictive representation for a supervised task. 

A popular flavor of graph neural network consists of stacked layers of operators  composed of learned transformations and neighbor aggregation. These  ``message-passing'' GNNs were inspired by spectral notions of graph convolution \citep{bruna2014spectral, defferrard2016convolutional, kipf2016semi}. Consider a graph $G$ with $n$ vertices and adjacency matrix $A\in R^{n\times n}$. A graph convolution $g\star x$ of node features $x$ by a filter $g$ is defined as a multiplication by $g$ in the graph Fourier basis, just as a standard convolution is a multiplication in Fourier space. The Fourier basis for a graph is defined as the eigenvectors $U$  of the normalized Laplacian, and can be thought of as a basis of functions of varying smoothness over the graph.
\begin{equation}
    g\star x = U g U^{T} x
\end{equation}

Any convolution $g$ can be approximated by a series of $k$-th order polynomials in the Laplacian, which depend on neighbors within a $k$-hop radius \citep{hammond2011wavelets}. By limiting this approximation to $k=1$, \cite{kipf2016semi} arrive at an operation that consists of multiplying node features by the normalized adjacency matrix, i.e. averaging each node's neighbor features. Such an operation can be viewed as a graph convolution by a particular smoothing kernel. A Graph Convolutional Network (GCN) is constructed by stacking multiple layers, each with a neighbor averaging step followed by a linear transformation. Many variants of this approach of stacked message-passing layers have since been proposed with different aggregation functions and for different applications \citep{Velickovic2018GAT, xu2018GIN, GraphSAGE, schlichtkrull2018modeling}.

Early GNN work focused on tasks with small graphs (thousands of nodes), and it's not straightforward to scale these methods to large-scale graphs. Applying neighbor aggregation by directly multiplying node features by the sparse adjacency matrix at each training step is computationally expensive and does not permit minibatch training. On the other hand, applying a GCN for a minibatch of labeled nodes requires aggregation over a receptive field (neighborhood) of diameter $d$ equal to the GCN depth, which can grow exponentially in $d$. Recent work in scaling GNNs to very large graphs have focused on training the GNN on sampled subsets of neighbors or subgraphs to allevate the computation and memory cost \citep{GraphSAGE,chen2017stochastic,zou2019layersampling,zeng2019graphsaint,chiang2019cluster}. 

\begin{wrapfigure}{R}{0.4\textwidth}
    \centering
    \includegraphics[width=\linewidth]{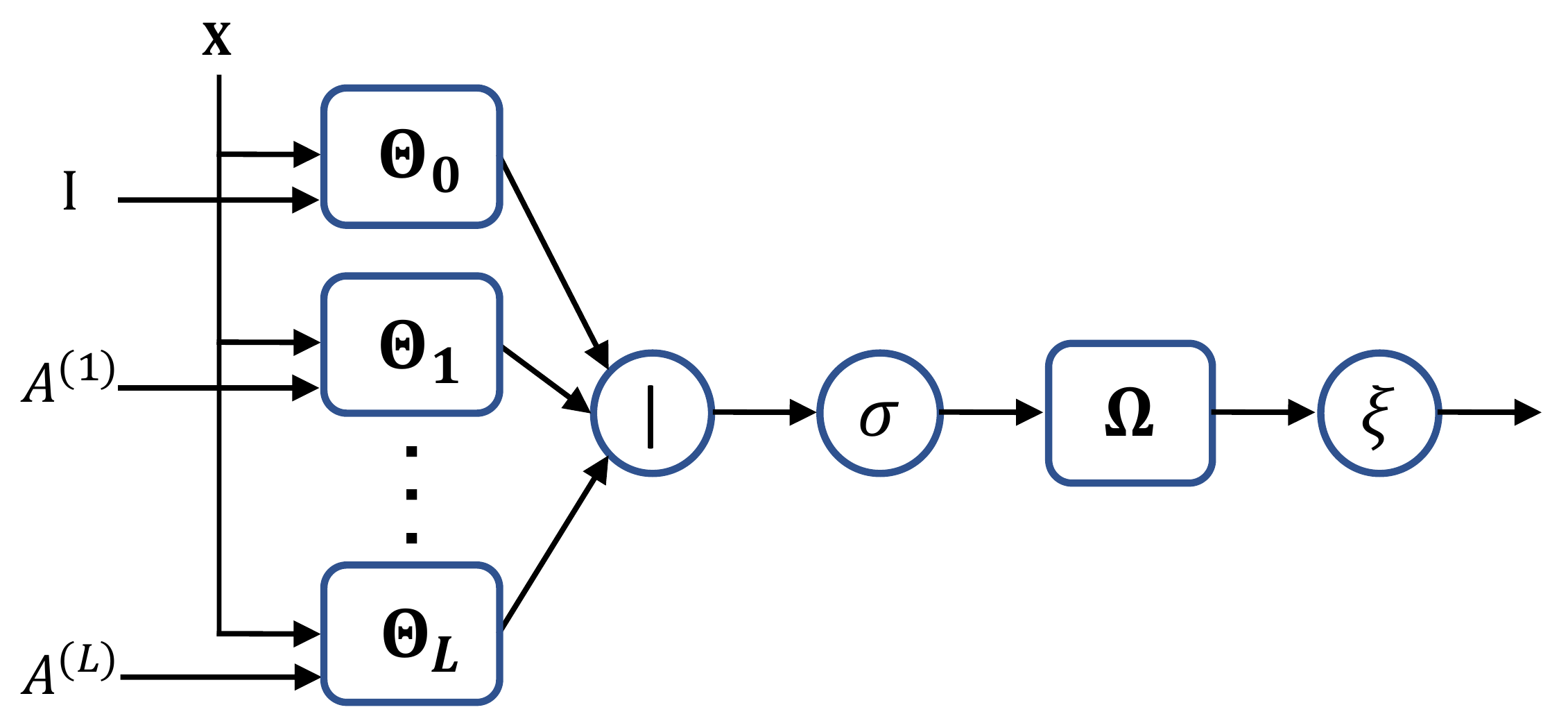}
    \caption{\small SIGN Model. $A^{(l)}$ is the $l$-th power of adjacency matrix $A$. $\Theta$ and $\Omega$ are transformation parameters in MLP, and $\sigma$ and $\xi$ are non-linear operators.}
    \label{fig:sign}
\end{wrapfigure}

\cite{rossi2020sign} proposed a different approach to scaling GCNs, 
called SIGN: As is shown in Figure \ref{fig:sign}, by eliding all learned parameters from intermediate layers, the GNN graph aggregation steps can be pre-computed as iterated neighbor feature averages, and model training consists of training an MLP on these neighbor-averaged features.  On benchmark tasks on large graphs, they observed that SIGN achieved similar accuracy to state-of-the-art GNNs. The success of SIGN suggests that GNNs are primarily using the graph to ``smooth'' node features over local neighborhoods rather than learning non-linear feature hierarchies. Similar hypotheses have been argued in several other recent works \citep{wu2019SGC, nt2019gfNN, chen2019powerful}\footnote{SIGN primarily differs from these other proposed methods by concatenating neighbor-aggregated features from different numbers of hops of feature aggregation. This addresses the need to balance the benefits of feature smoothing from large neighborhoods with the risk of ``oversmoothing'' the features and losing local neighborhood information when GNNs are too deep \citep{li2018deeper}, allowing the classifier to learn a balance between features from different GNN depths.}

Standard GCNs extend naturally to heterogeneous (aka relational) graphs by applying relation-specific learned transformations \citep{schlichtkrull2018modeling}. There have also been a number of GNN variants specialized to heterogeneous graphs. HetGNN~\citep{zhang2019heterogeneous} performs fixed-size random walk
on the graph and encodes heterogeneous content with RNNs. Heterogeneous
Attention Network~\citep{wang2019heterogeneous} generalizes neighborhood of
nodes based on semantic patterns (called metapaths) and extends
GAT~\citep{Velickovic2018GAT} with a semantic attention mechanism. The Heterogeneous
Graph Transformer~\citep{hu2020heterogeneous} uses an attention mechanism that conditions on
node and edge types, and introduces relative temporal encoding to handle dynamic
graphs. These models inherit the scaling limitation of GNN and are expensive to train on large graphs. Therefore, it is of practical importance to generalize the computationally much simpler SIGN model to heterogeneous graphs.

\section{\approachfull for Heterogeneous Graphs}
The challenge with adapting SIGN to heterogeneous graphs is how to
incorporate the information about different node and edge types.
Relational GNNs like R-GCN naturally handle heterogeneous graphs by learning
different transformations for each relation type \citep{schlichtkrull2018modeling}, but SIGN elides these learned transformations. While one can naively apply SIGN to heterogeneous graphs by ignoring entity and relation types, doing so results in poor task performance (Table~\ref{tab:res}).

In this section, we propose {\em Neighbor Averaging over Relation Subgraphs} (NARS). The key idea of NARS is to utilize entity and relation information by
repeatedly sampling subsets of relation types and building subgraphs consisting
only of edges of these types, which we call relation subgraphs. We then
perform neighbor averaging on these relation subgraphs and aggregate features
with a learned 1-D convolution. The resulting features are then used to train an
MLP, as in SIGN. 

We take inspiration from the notion of ``metapaths'' proposed in
\cite{dong2017metapath2vec} and used by several recent heterogeneous GNN
approaches~\citep{wang2019heterogeneous}. A metapath is a sequence of relation
types that describes a semantic relationship among different types of nodes;
for example, one metapath in an academic graph might be ``venue - paper -
author'', which could represent ``venues of papers with the same author as the
target paper node''. Information passed and aggregated along a metapath is
expected to be semantically meaningful. In previous work like
HAN~\citep{wang2019heterogeneous}, features from different metapaths
are later aggregated to capture different neighborhood information for each
node.

In prior work, relevant metapaths were manually specified as input
\citep{dong2017metapath2vec,wang2019heterogeneous}, but we hypothesized that the
same information could be captured by randomly sampling metapaths. However,
sampling individual metapaths doesn't scale well to graphs with many edge types:
for a graph with $M$ edge types, it would require $O(M)$ metapaths to even
sample each edge type in a single metapath. 
As a result, one might need to sample a large set of metapaths to obtain good prediction results, and in practice we obtained poor task performance by sampling metapaths.

We observe that metapaths are an instance of a more general class of aggregation procedures: those that aggregate at each GNN layer over a \textit{subset} of relation types rather than a single relation type. We consider a different procedure in that class: sampling a subset of relation types (uniformly from the power set of relation types), and using this subset for all aggregation layers. This procedure amounts to randomly selecting $K$ \textit{relation subgraphs} each specified by a subset of relations, and performing $L$-hop neighbor aggregation across each of these subgraphs. We found that this strategy led to strong task performance (\S\ref{sec:eval}), but it's possible that other aggregation strategies could perform even better.

Given a heterogeneous graph $G$ and its edge relation type set $\mathcal{R}$,
our proposed method first samples $K$ unique subsets from $\mathcal{R}$. Then for each sampled subset $R_i \subseteq \mathcal{R}$, we generate a relation subgraph $G_i$ from $G$ in which only edges whose type belongs to ${R_i}$ are kept. We treat $G_i$ as a homogeneous graph,
and perform neighbor aggregation to generate $L$-hop neighbor features for each
node.

Let $H_{v,0}$ be the input features (of dimension $D$) for node $v$. For each subgraph $G_i$, the $l$-th hop features $H_{v,l}^i$ are computed as
\begin{equation}
    H_{v,l}^i = \sum\limits_{u\in N_i(v)}\frac{1}{|N_i(v)|}H_{u,l-1}^i
\end{equation}
where $N_i(v)$ is the set of neighbors of node $v$ in $G_i$. 

\subsection{Aggregating SIGN features from sampled subgraphs}

For each layer $l$, we let the model adaptively learn which relation-subgraph
features to use by aggregating features from different subgraphs $G_i$ with a
learnable 1-D convolution. The aggregated $l$-hop features across all subgraphs are calculated as
\begin{equation}
\label{eq:agg_feat}
H_{v,l}^{agg} = \sum\limits_{i=1}^K a_{i,l} \cdot H_{v,l}^i
\end{equation}
\begin{wraptable}{r}{60mm}
\centering
\begin{tabular}{l l}
\hline
\bf Model & \bf \# parameters \\
\hline
SIGN & $D^2 L$ \\
\approach (concat) & $D^2 L K$ \\
\approach (1D conv) & $D^2 L + D L K$ \\
\hline
\end{tabular}
\caption{\small Number of parameters (equivalently, FLOPs per prediction) in vanilla SIGN and \approach. If features are concatenated in \approach, the number of parameters grows multiplicatively, but this is resolved by using 1D convolution to reduce input dimension to the classifier.\label{tab:param}}
\vspace{-0.5cm}
\end{wraptable}
where $H^i$ is the neighbor averaging features on subgraph $G_i$ and
$a_{i,l}$ is a learned vector of length equal to the feature dimension $D$. In total, $\bf a$ is a tensor of learned coefficients of size $K \times L \times D$. 

We use a 1D convolution to reduce the number of input parameters to the subsequent SIGN classifier, avoiding a multiplicative increase in the number of model parameters by a factor of $K$, as shown in Table~\ref{tab:param}. Having fewer parameters reduces the cost of computation and memory and is less prone to overfitting. 

A classifier is then trained on the aggregated node features to predict task labels, using the MLP architecture described in SIGN \citep{rossi2020sign}.

\subsection{Node types without input features}
One consideration for learning on heterogeneous graphs is that input features are not always provided for all node types. Take the OAG academic graph as an example: in prior work, paper nodes were featurized with language embeddings for the paper title using a pretrained XLNet model~\citep{yang2019xlnet}. But for other node types like author, field, venue, node features were not provided, so any features must be inferred from the graph.

There are different ways to handle these ``featureless'' node types, and the best
approach might be dependent on the dataset and task. On the tasks we examined, we found it helpful to use relational graph embeddings \citep{bordes2013translating} trained on the heterogeneous graph as the initial features for nodes without provided input features. Note that these graph embeddings do not make use of the input features provided for the papers.

We compare the effectiveness of using different types of features for nodes that don't have intrinsic (i.e. content) features in Section~\ref{sec:eval:input_feat}.

\section{Memory footprint optimization}
\label{sec:mem}

\approach allows trainable aggregation of features from sampled subgraphs.
However, even though the model is trained in minibatch fashion on GPU, we
have to precompute and store all the subgraph-aggregated features in CPU memory. The amount of memory required to store these pre-computed features is $O(NLDK)$, which is $K$ times more than SIGN. So for large heterogeneous graphs with many edge types, there is a tradeoff between sampling more subgraphs in order to capture semantically meaningful relations, and limiting the number of subgraphs to conserve memory during training.

To address this issue, we propose to divide the training into multiple stages.
In each stage which lasts several epochs, we train the model with a subset of the sampled subgraphs. Concretely speaking, we approximate Equation~\eqref{eq:agg_feat} with the following equation:
\begin{equation}
    H_{v,l}^{(t)}=\sum\limits_{G_i\in S^{(t)}\subseteq S}b_{i,l}\cdot H_{v,l}^i+\alpha H_{v,l}^{(t-1)}
\label{eq:part_agg_feat}
\end{equation}
In this equation, $S^{(t)}$ is the randomly sampled subset of the $K$ subgraphs used in stage $t$, and $H^{(t)}$ is the approximation of $H^{agg}$ at stage $t$.

Equation~\eqref{eq:part_agg_feat} is equivalent to setting
$a_{i,l}^{(t)}=b_{i,l}+\alpha a_{i,l}^{(t-1)}$ for $G_i\in S^{(t)}$ and
$a_{i,l}^{(t+1)}=\alpha a_{i,l}^{(t)}$ for $G_i\notin S^{(t)}$
in Equation~\eqref{eq:agg_feat}, so the values of $\bf a$ can be updated after each stage. The approximated $H^{(t)}$ is used as the input to the classifier during end-to-end training across all stages. The detailed training process is shown in Algorithm~\ref{alg:part_train} in the Appendix.

Our sub-sampling approach reduces memory usage during training to
$O(NLD|S^{(t)}|)$, and in practice, we found that even $|S^{(t)}|=1$ produces decent results and outperforms the accuracy of current SOTA models (see \S\ref{sec:eval:mem}).
One thing worth pointing out is that even though this memory optimization
requires regenerating features for $|S^{(t)}|$ subgraphs every a few epochs, the
generation can be done on CPU and can completely overlap with GPU training, and hence does not slow down training.

For inference, since all model parameters in Equation~\eqref{eq:agg_feat} are fixed,
instead of using our proposed memory optimization method, we simply generate
features for each sampled subgraph and compute the aggregation in place. The memory overhead for inference is therefore also only $O(NDL)$.

\section{Evaluation}
\label{sec:eval}
In this section, we evaluate \approach on several popular heterogeneous graph
datasets and compare with state-of-the-art models. We also investigate
the effect of different ways to featurize nodes without input features, and how the memory optimization described in \S\ref{sec:mem}
 affects prediction accuracy.

\subsection{Experimental Setup}
\begin{table}
\tabcolsep=0.15cm
\scriptsize
\begin{tabular}{c|rrrcc|rrccc}
\hline
\multirow{2}{*}{Dataset} & \multicolumn{5}{c|}{Node Types}                                                                                                           & \multicolumn{5}{c}{Edge Types}                                                                                                                           \\ \cline{2-11} 
                         & \multicolumn{1}{c}{\# P} & \multicolumn{1}{c}{\# A} & \multicolumn{1}{c}{\# F} & \# I                       & \# V                        & \multicolumn{1}{c}{\# P-A} & \multicolumn{1}{c}{\# P-F} & \# P-P                        & \#A-I                         & \# P-V                         \\ \hline
ACM                      & 4,025                    & 17,431                   & 73                       & $-$                        & $-$                         & 13,407                     & 4,025                      & $-$                           & $-$                           & $-$                            \\
OGB-MAG                  & 736,389                  & 1,134,649                & 59,965                   & \multicolumn{1}{r}{8,740}  & $-$                         & 7,145,660                  & 7,505,078                  & \multicolumn{1}{r}{5,416,271} & \multicolumn{1}{r}{1,043,998} & $-$                            \\
OAG (CS)                 & 5,597,605                & 5,985,759                & 119,537                  & \multicolumn{1}{r}{27,433} & \multicolumn{1}{r|}{16,931} & 15,571,614                 & 47,462,559                 & \multicolumn{1}{r}{5,597,606} & \multicolumn{1}{r}{7,190,480} & \multicolumn{1}{r}{31,441,552} \\ \hline
\end{tabular}
\caption{\small Statistics of three academic graph datasets. The node types are: Paper
    (P), Author (A), Field (F), Institute (I) and Venue (V)}.
\label{tab:graph_stat}
\vspace{-0.5cm}
\end{table}
\paragraph{Datasets \& Tasks}
We evaluate our model using node prediction on three popular academic
graph datasets: ACM~\citep{wang2019heterogeneous}, OGB-MAG~\citep{hu2020open}, and OAG~\citep{sinha2015overview,tang2008arnetminer,zhang2019oag}. 
The tasks involve predicting a paper's category, its
publishing venue, or its research field on these three
datasets.
We summarize the statistics of the datasets in
Table~\ref{tab:graph_stat} and put the details of each dataset in Appendix
(\ref{sec:app:dataset}).

\begin{table}
\small
\resizebox{\columnwidth}{!}{
\begin{tabular}{c|cccc|rrrrr}
\hline
\multicolumn{1}{c|}{\multirow{2}{*}{Dataset}} & \multicolumn{4}{c|}{Hyper-parameters}                                                             & \multicolumn{5}{c}{\# Model Parameters}                                                                                                                                                   \\ \cline{2-10} 
\multicolumn{1}{l|}{}                         & \multicolumn{1}{c}{\# hidden} & \multicolumn{1}{c}{\# layers (L)} & \multicolumn{1}{c}{\# subgraphs (K)}  & \multicolumn{1}{c|}{TransE size}    & \multicolumn{1}{c}{\approach (Ours)}  & \multicolumn{1}{c}{SIGN} & \multicolumn{1}{c}{R-GCN}  & \multicolumn{1}{c}{HAN}   & \multicolumn{1}{c}{HGT} \\ \hline
ACM                                           & 64                            & 2                             & 2                                 & 128                                 & 0.40M                                 & 0.39M                    & 0.14M                      & 0.25M                     & 0.26M                   \\
OGB-MAG                                       & 512                           & 5                             & 8                                 & 256                                 & 4.13M                                 & 4.12M                    & 9.18M            & \multicolumn{1}{c}{--}    & 26.88M                  \\
OAG-Venue                                     & 256                           & 3                             & 8                                 & 400                                 & 2.24M                                 & 2.21M                    & 40.60M           & \multicolumn{1}{c}{--}             & 8.26M                   \\
OAG-L1-Field                                  & 256                           & 3                             & 8                                 & 400                                 & 1.41M                                 & 1.38M                    & 11.64M           & \multicolumn{1}{c}{--}              & 7.43M                   \\ \hline
\end{tabular}
}
    \caption{\small Training hyper-parameters \& number of parameters for each
    model. For hyper-parameters \# hidden and \# layers, we adopt values from
    HAN for ACM, and values from HGT for OGB-MAG and OAG. \# subgraphs sampled
    for \approach is picked to be small while producing good and stable results.
    TransE size is selected based on the size of the graph.}
\label{tab:params}
\end{table}
\paragraph{Baselines \& Metrics}

We compare \approach with four baseline models:
R-GCN~\citep{schlichtkrull2018modeling}, HAN~\citep{wang2019heterogeneous},
HGT~\citep{hu2020heterogeneous}, and SIGN~\citep{rossi2020sign}. The first three
models are designed for heterogeneous scenarios and can be naturally applied on
all datasets. The vanilla SIGN model, however, is for the homogeneous graph setting.
Therefore, we ignore the node types and edge types when training SIGN. 
We use the best implementation we can find for each baseline model: for HGT, we use the authors' \href{https://github.com/acbull/pyHGT}{original implementation}. For \href{https://github.com/dmlc/dgl/tree/master/examples/pytorch/sign}{SIGN} and \href{https://github.com/dmlc/dgl/tree/master/examples/pytorch/han}{HAN}, we use the implementation by Deep Graph Library \citep{wang2019dgl}. For  \href{https://github.com/snap-stanford/ogb/tree/master/examples/nodeproppred/mag}{R-GCN}, we use the implementation by PyTorch Geometric~\citep{Fey/Lenssen/2019}.

We report test accuracy (micro-F1) for the ACM and
OGB-MAG datasets and NDCG and MRR for the OAG dataset, using the model from the epoch that performs best on the validation set. 
For each experiment, we run five replicates and report the
average and standard deviation across replicates.

Training GNN baselines on the OGB-MAG and OAG datasets is intractable due to memory usage, unless sampling is used. We adopt neighbor sampling \citep{GraphSAGE} for R-GCN and HGT’s custom sampling method, but were unable to evaluate HAN on these datasets, since it is unclear how to train this model with sampling while following metapath constraints.

\paragraph{Training settings}
We train \approach with learning rate 0.001, dropout rate 0.5 and the Adam optimizer. For
nodes that don't have input features, we use pre-trained TransE embeddings~\citep{bordes2013translating} as
input.
During the preprocessing phase, we sample $K$ subgraphs and pre-compute the aggregated features of $L$ hop neighbors for each node. Table~\ref{tab:params} shows the training
hyper-parameters for our model, as well as a comparison of model sizes (in terms of the number of  parameters) across ours and various baseline models. Though we try to use the same
hyper-parameters for all models, we note that our extension to
SIGN results in a simple model and in most cases has significantly fewer training
parameters.

Detailed implementation and configuration can be found in our open-sourced
GitHub repository\footnote{\url{https://github.com/facebookresearch/NARS}}.

\subsection{Results}
\begin{table}
\resizebox{\columnwidth}{!}{
\begin{tabular}{c|l|l|ll|ll}
\hline
Dataset          & ACM                       & OGB-MAG                   & \multicolumn{2}{c|}{OAG-Venue}                    & \multicolumn{2}{c}{OAG-L1-Field}                \\ \hline
Metric           & \cellalign{c|}{Accuracy}  & \cellalign{c|}{Accuracy}  & \cellalign{c}{NDCG}      & \cellalign{c|}{MRR}    & \cellalign{c}{NDCG}    & \cellalign{c}{MRR}     \\ \hline
R-GCN            & \textbf{.930$\pm$.002}    & .500$\pm$.001   & .481$\pm$.004  & .302$\pm$.005 & .852$\pm$.002   & .843$\pm$.002   \\
HAN              & .922$\pm$.002             & \multicolumn{1}{c|}{--}   & \multicolumn{1}{c}{--}  & \multicolumn{1}{c|}{--}& \multicolumn{1}{c}{--}  & \multicolumn{1}{c}{--}   \\
HGT\tablefootnote{We improved the reported performance of HGT by 5\% by sampling 10 times for target nodes and using the average of predictions, to reduce sampling variance.}              & .919$\pm$.003             & .498$\pm$.001             & .498$\pm$.014            & .322$\pm$.014          & \textbf{.868$\pm$.002} & .849$\pm$.003          \\
SIGN             & .919$\pm$.001             & .481$\pm$.001             & .506$\pm$.001            & .327$\pm$.001          & .839$\pm$.001          & .826$\pm$.003          \\
\approach (Ours) & \textbf{.931$\pm$.004}    & \textbf{.521$\pm$.004}    & \textbf{.520$\pm$.003}   & \textbf{.342$\pm$.003} & \textbf{.868$\pm$.001} & \textbf{.857$\pm$.003} \\ 
\hline
\end{tabular}
}
\caption{\small Performance of \approach vs. baseline models on different datasets and
    tasks. All numbers are average and standard deviation over 5 runs. Bold
    numbers indicate best model(s).}
\label{tab:res}
\end{table}

Table~\ref{tab:res} summarizes the performance of \approach and baseline models
on different datasets and tasks. \approach outperforms all baseline methods on all datasets and node prediction tasks that we evaluated.
In particular, \approach improves test performance by up to 4\% compared to the current
state-of-the-art model (HGT) on large datasets like OAG and MAG. Compared to naively applying SIGN by treating the graph as homogeneous, \approach improves
performance by up to 8\%. 
It is quite surprising that \approach exceeds the performance of existing
state-of-the-art GNN methods since \approach is computationally
much simpler than the baselines; like SIGN, it does not learn any feature hierarchies over graph neighborhoods, but merely learns a classifier on graph-smoothed features.

Figure~\ref{fig:speed} compares the
training speed of \approach and public implementations of baseline methods on the OGB-MAG
dataset. We did not include the time for pre-training TransE
embeddings (40 min) since all models benefit from using them
(\S\ref{sec:eval:input_feat}).
In Figure~\ref{fig:speed}, the solid blue line labelled \approach (CPU features) 
refers to an implementation that stores all pre-computed features on the CPU and transfers them to the GPU on demand during each mini-batch training. This \approach implementation not only achieves superior accuracy, but also is
substantially faster to train than competing approaches. We can further improve \approach training speed by storing pre-computed features on the GPU if space permits, as shown 
by the dotted blue line labeled \approach (GPU features).

\subsection{Effect of number of subgraphs sampled}
\begin{figure}
    \centering
    \begin{minipage}[t]{0.48\textwidth}
        \centering
        \includegraphics[width=1.04\linewidth]{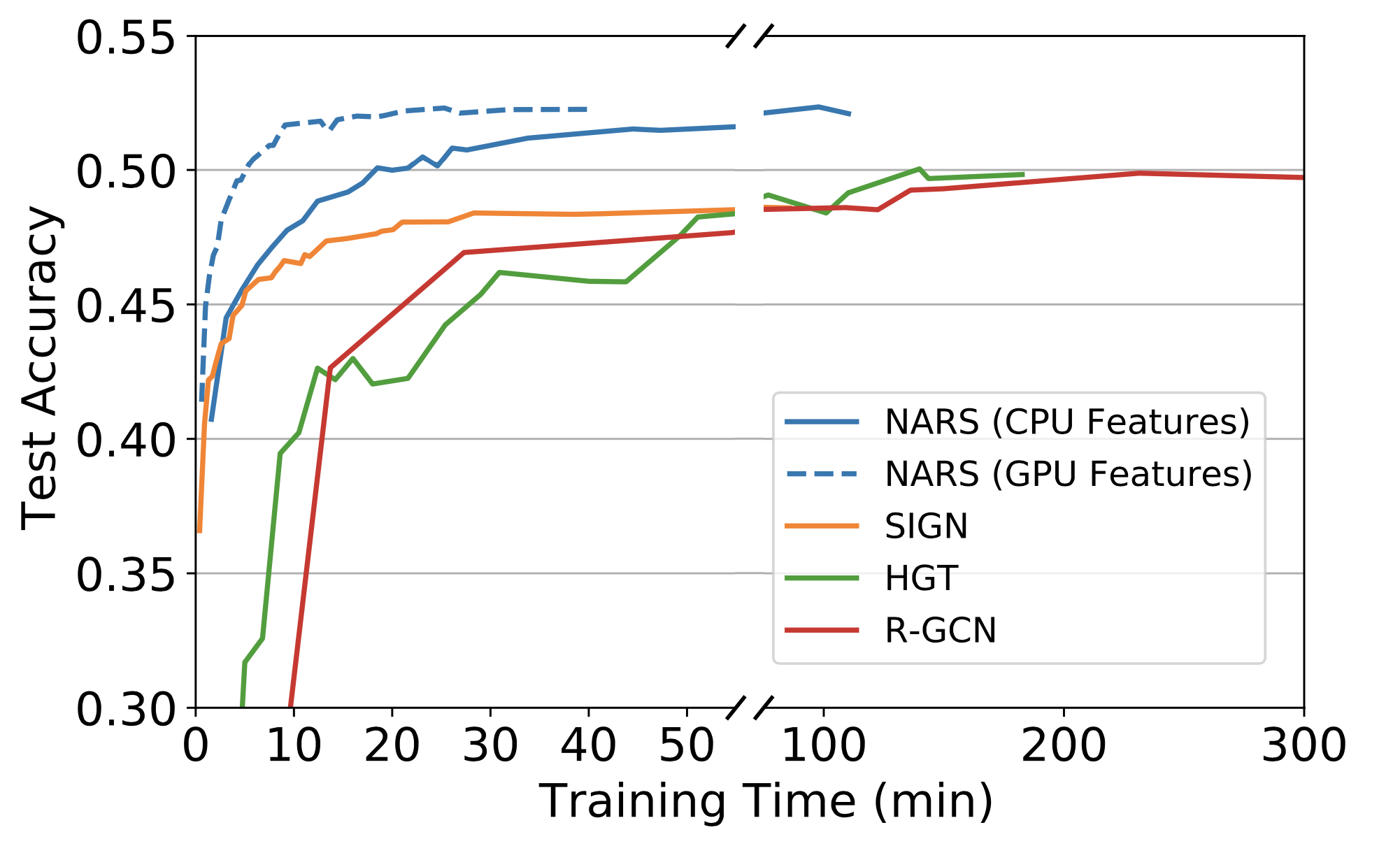}
        \caption{\small Training speed of \approach (sampling 8 subgraphs) and baseline models on the OGB-MAG dataset. \approach leads to higher final accuracy in less training time. NARS training is much faster when features can be stored on GPU (dashed line).}
        \label{fig:speed}
    \end{minipage}
    \hfill
    \begin{minipage}[t]{0.48\textwidth}
        \centering
        \includegraphics[width=\linewidth]{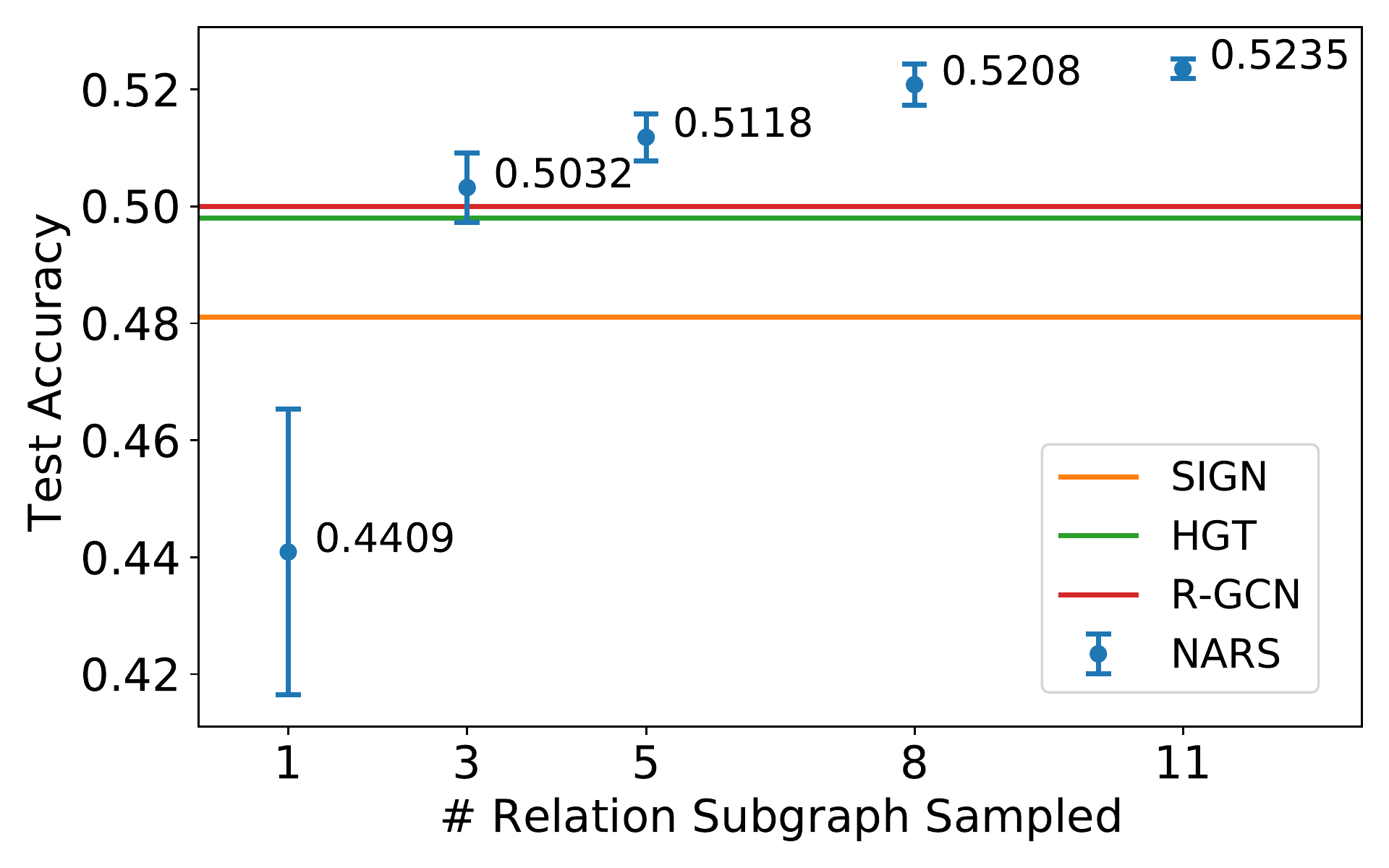}
        \caption{\small Accuracy of sampling different numbers of relation subgraphs on the OGB-MAG
        dataset. Sampling more relation subgraphs improves test accuracy while reducing variance across replicates.}
        \label{fig:num_sample}
    \end{minipage}
\vspace{-0.5cm}
\end{figure}

We have shown in Table~\ref{tab:res} that sampling relation subgraphs
produces better results than treating the graph as homogeneous. Now we use the OGB-MAG dataset to perform an ablation study on how the number
of sampled subgraphs affects the performance.

OGB-MAG dataset has 4 edge relation types, hence the edge relation set $R$ has
in total 15 non-empty subsets. Among the 15 subsets, only 11 are valid (others
don't touch paper nodes for prediction or are disconnected). In
Figure~\ref{fig:num_sample}, we randomly sample subsets of size 1, 3, 5, 8 from
$R$ which contains the 11 valid subgraphs. For each point in the figure, we
randomly sample 3 different subsets and report the average and standard
deviation.

As shown in Figure~\ref{fig:num_sample}, the average accuracy improves while the variance
decreases as the number of sampled subgraphs increases, which is expected since
sampling more reduces sampling variance and increase the chance to cover
relation subgraphs that are more semantically meaningful. On the OGB-MAG
dataset, sampling $\geq 5$ subgraphs always outperforms current
state-of-the-art models like HGT (green line) and R-GCN (red line).

\subsection{Effect of different strategies to handle featureless nodes}
\label{sec:eval:input_feat}
\begin{table}
    \tabcolsep=0.38cm
    \resizebox{\columnwidth}{!}{
    \small
    \begin{tabular}{c|cccc}
        \hline
                      & Padding Zeros   & Average Neighbors & Metapath2vec Embs    & TransE Embs \\
        \hline
        \approach     & .4462$\pm$.0022 & .4427$\pm$.0018   & .5187$\pm$.0011 & .5214$\pm$.0016     \\
        SIGN          & .4007$\pm$.0010 & .4028$\pm$.0021   & .4685$\pm$.0011 & .4810$\pm$.0009     \\
        HGT           & .4597$\pm$.0027 & .4891$\pm$.0021   & .4962$\pm$.0026 & .4982$\pm$.0013     \\
        R-GCN         & .4811$\pm$.0028 & .4707$\pm$.0033   & .5013$\pm$.0019 & .5001$\pm$.0005     \\
        \hline
    \end{tabular}
    }
    \caption{\small Comparison of different ways to featurize nodes
    with no input features on the OGB-MAG dataset. All models achieve their best performance with pre-trained TransE graph embedding features. Featurization is especially important for neighbor-averaging approaches (SIGN and \approach).}
    \label{tab:input_feat}
    \vspace{-0.3cm}
\end{table}
In this section, we compare how different ways of featurizing nodes that don't
have input features affect the performance of \approach on OGB-MAG dataset.
In the dataset, only paper nodes have input language features generated using
word2vec, and all other node types are not associated with any input
features.

We tried four different ways to generate features for the featureless nodes:
1) padding zero; 2) taking the average of features from neighboring papers
nodes; 3) using pre-trained Metapath2vec embedding; 4) using pre-trained
TransE relational graph embeddings with L2 norm. We followed the same hyper-parameters listed in Table~\ref{tab:params}.

As shown in Table~\ref{tab:input_feat}, unsupervised graph
embedding methods greatly improve model accuracy, especially for SIGN and NARS. We use TransE embeddings in the rest of this section because it achieves the best accuracy, and because metapath2vec embeddings require manually specifying metapaths, which we seek to avoid.

\subsection{Effect of training with subset of sampled subgraphs}
\label{sec:eval:mem}
\begin{wrapfigure}{R}{0.5\textwidth}
\vspace{-0.5cm}
    \centering
    \includegraphics[width=\linewidth]{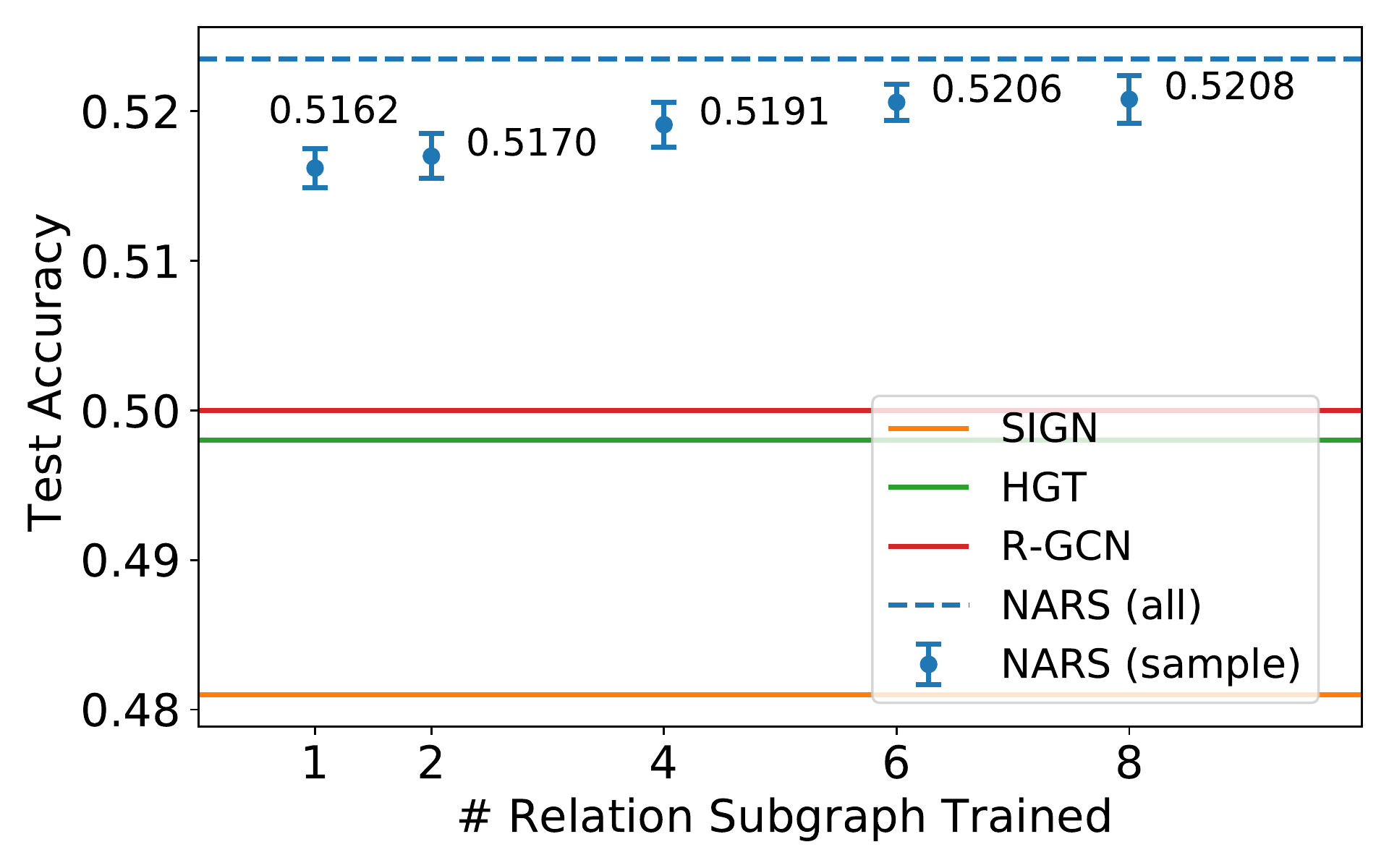}
    \caption{\small Training with different numbers of subgraphs in each stage (\S\ref{sec:mem})
    on OGB-MAG dataset, to improve memory efficiency. Good test accuracy is achieved with even a single subgraph sampled per training stage.}
    \label{fig:part_train}
    \vspace{-1cm}
\end{wrapfigure}
In section~\ref{sec:mem}, we proposed to train with a sub-sampled set from all
$K$ sampled relation subgraphs to reduce CPU memory usage. In
Figure~\ref{fig:part_train}, we vary the number of relation subgraphs in the randomly
sampled subset in each stage on OGB-MAG dataset to see how this approach affects
accuracy. The blue dashed line at the top of the figure is the accuracy
for training with all 11 valid relation subgraphs. The blue point with error bar
represents the average and standard deviation for sampling a certain number of
subgraphs from the 11 valid relation subgraphs in each stage. 
Performance improves when more subgraphs are sampled in each stage, but sampling a single subgraph in each stage leads to good performance with small variance, outperforming existing models.

\section{Conclusion}

Simplified GNNs like SIGN that do not require learned aggregation are a promising new class of models for graph learning due to their simplicity, scalability, and interpretability. We present \approachfull, a novel GNN architecture for heterogeneous graphs that learns a classifier based on a combination of neighbor-averaged features for random subgraphs specified by a subset of relation types. \approach beats state-of-the-art task performance on several benchmarks despite its simpler and more scalable approach.

This work provides further evidence that current GNNs might not learn meaningful feature hierarchies on benchmark datasets, but are primarily functioning by graph feature smoothing. Future advances in GNN modeling may realize the benefits of ``deep'' learning for these tasks, or it may be that this modeling is not necessary for many important graph learning tasks.

One remaining limitation of \approach is that it doesn't explicitly handle heterogeneous feature types, averaging together features of different types (e.g. language features vs. graph features). While this performs adequately on benchmark datasets, it is an unsatisfying approach for industry graph datasets with potentially dozens or hundreds of distinct entity types, each with their own distinct features. In these situations, existing GNNs (e.g. R-GCN) may be more appropriate, and neighbor averaging approaches suitable for this situation are an area for future work.

\bibliography{refs}
\bibliographystyle{iclr2021_conference}

\newpage
\appendix
\section{Appendix}

\subsection{Algorithm to train \approach}
\label{sec:app:train_alg}
Algorithm~\ref{alg:full_train} demonstrates the detailed training algorithm for \approachfull.
\begin{algorithm}
\label{alg:full_train}
\KwIn{$G$ a heterogeneous graph}
\KwIn{$\mathcal{R}$ the set of relations in $G$}
\KwIn{$K$ number of relation subgraphs to sample}
\KwIn{$L$ number of layers in MLP model}
\KwIn{$H_{input}$ node input features}

\SetKwProg{Def}{def}{:}{}
\tcc{Generate multi-hop neighborhood-averaged features}
\Def{GenNeighborFeature($G$)}{
    \tcp{Build adjacency matrix for $G$ regardless of edge types}
    $A \gets BuildAdjacencyMatrix(G)$\;
    \tcp{Divide nonzeros in each row by node in-degree}
    $W \gets RowNormalize(A)$\;
    $H_0 \gets H_{input}$\;
    \For{$l$ $\gets 1$ \textbf{to} $L$}{
        $H_{l} \gets WH_{l - 1}$\;
    }
    \KwRet $H_0, H_1, \dots, H_{L}$\;
}
\tcc{Sample $K$ unique relation subgraphs from $G$ and $\mathcal{R}$}
\Def{SampleRelationSubgraph($G$, $\mathcal{R}, K$)}{
    $P \gets GetPowerSet(\mathcal{R})$\;
    $S \gets SampleWithoutReplacement(P, K)$\;
    $E \gets GetEdgeSet(G)$\;
    \For{$R_i \in S$}{
        $E_i \gets \emptyset$\;
        \For{$e \in E$}{
            \If{$e.type \in R_i$}{
                $E_i \gets E_i + \{e\}$\;
            }
        }
        $G_i \gets BuildSubgraphFromEdgeSet(E_i)$\;
    }
    \KwRet $G_0, G_1,\dots, G_{K-1}$\;
}
\tcc{Prepreocssing}
$S\gets SampleRelationSubgraph(G, \mathcal{R}, K)$\;
\For{$G_i\in S$}{
    $H_0^i,H_1^i,\dots,H_{L}^i\gets GenNeighborFeature(G_i)$
}
\tcc{Training}
\tcp{Any MLP classfier, SIGN model for example}
$model \gets MLPModel()$\;
\For{epoch $\gets 1$ \textbf{to} MAX\_EPOCH} {
    \tcp{Compute Equation~\ref{eq:agg_feat}}
    \For {$l\gets 0$ \textbf{to} $L$}{
        $H_l\gets\sum\limits_{i=0}^{K}a_{l}^iH_l^i$\;
    }
    $loss\gets model.forward(\{H_l\})$\;
    $loss.backward()$\;
    $gradient\_update(\{a_l^i\}, model)$\;
}
\caption{\approachfull}
\end{algorithm}

\subsection{Algorithm to train with reduced memory usage}
Algorithm~\ref{alg:part_train} demonstrates details about how to further sub-sample $p$ subgraphs in each stage from the set of relation subgraph $S$ and use Equation~\eqref{eq:part_agg_feat} to train the model. In Algorithm~\ref{alg:part_train}, we omit the layer number $l$ to make it more concise.
\begin{algorithm}
\KwIn{$S$ a set of $K$ sampled subgraphs}
\KwIn{$T$ number of epochs in each stage}
\KwIn{$p$ number of subgraphs used for training in each stage}
$model \gets MLPModel()$\;
$Uniform\_Random\_Init(\{a_k\})$\;
\tcp{Initialize history of aggregated features using Equation~\ref{eq:agg_feat}}
$H^{(0)} \gets 0$\;
\For{$G_i\in S$}{
    $H^i \gets GenNeighborFeature(G_i)$\;
    $H^{(0)} \gets H^{(0)} + a_{i} H^{i}$\;
}
\tcp{Sample a subset $S^{(0)}$ from $S$}
$S^{(0)} \gets SampleSubset(S, p)$\;
\For{$G_i\in S^{(0)}$}{
    \tcp{Generate neighbor feature for each sampled $G_i \in S^{(0)}$}
    $H^{i} \gets GenNeighborFeature(G_i)$\;
    \tcp{Initialize $b_{i}$ to $0$}
    $b_{i} \gets 0$\;
}
\tcp{Initialize $\alpha$ to $1$}
$\alpha\gets 1$\;
\tcp{Initialize stage $t$ to $1$}
$t\gets 1$\;
\For{epoch $\gets 1$ \textbf{to} MAX\_EPOCH} {
    \tcp{Compute approximated aggregation following Equation~\ref{eq:part_agg_feat}}
    $H^{(t)} \gets \alpha H^{(t-1)}$\;
    \For{$G_i\in S^{(t)}$}{
        $H^{(t)} \gets H^{(t)} + b_i H^{i}$\;
    }
    $loss \gets model(H^{(t)})$\;
    $loss.backward()$\;
    $gradient\_update(model, \{b_i\}, \alpha)$\;
    \If{epoch \% T $== 0$}{
        \tcp{Update history of aggregation}
        $H^{(t)} \gets \alpha H^{(t-1)}$\;
        \For{$G_i\in S^{(t)}$}{
            $H^{(t)} \gets H^{(t)} + b_{i} H^{i}$\;
        }
        \tcp{Update $a_{i}$}
        \For{$G_i\in S^{(t)}$}{
            $a_{i}\gets b_{i} + \alpha a_{i}$\;
        }
        \For{$G_i\notin S^{(t)}$}{
            $a_{i}\gets \alpha a_{i}$\;
        }
        \tcp{Increment stage $t$}
        $t\gets t+1$\;
        \tcp{Re-sample $S^{(t)}$ from $S$}
        $S^{(t)} \gets SampleSubset(S, p)$\;
        \For{$G_i\in S^{(t)}$}{
            \tcp{Generate neighbor feature for newly sampled $G_i \in S^{(t)}$}
            $H^{i} \gets GenNeighborFeature(G_i)$\;
            \tcp{Re-initialize $b_{i}$ to $0$}
            $b_{i} \gets 0$\;
        }
        \tcp{Re-initialize $\alpha$ to $1$}
        $\alpha\gets 1$\;
    }
}

\caption{Train with subset of sampled subgraph}
\label{alg:part_train}
\end{algorithm}

\subsection{Datasets}
\label{sec:app:dataset}
\paragraph{ACM}
We use the ACM dataset from Heterogeneous Graph Attention
Network~\citep{wang2019heterogeneous}, which is an academic graph extracted from
papers published in ACM conferences (KDD, SIGMOD, SIGCOMM, MobiCOMM, and VLDB).
The HAN author-provided version removed the field and author node types.
Therefore, we used the version re-constructed by DGL~\citep{wang2019dgl}. In
this dataset, only paper nodes have Bag-of-Words features. Papers are divided
into three classes and the task is to predict the class for each paper.
\paragraph{OGB-MAG}
Open Graph Benchmark~\citep{hu2020open} is an effort to build a standard, large
and challenging benchmark for graph learning, which contains a large collection
of datasets that cover important tasks on graphs and a wide range of domains.
Leaderboards are set up for each dataset and performance of
state-of-the-art models is listed with open-sourced implementation to reproduce
results. We evaluate our approach on the MAG benchmark from OGB node prediction
category, which is a heterogeneous network extracted from Microsoft Academic
Graph (MAG). The papers are published on 349 different venues and they come with
Word2Vec features. The task here is to predict the publishing venue for each paper.
\paragraph{OAG}
Open Academic Graph~\citep{sinha2015overview,tang2008arnetminer,zhang2019oag}
is the largest public academic graph with more than 178
million nodes and 2 billion edges. We use the pre-processed CS domain
component\footnote{HGT authors only shared the CS component. Also they performed
a filter step to make the graph denser. After the filtering, there are 544244
papers, 45717 fields, 510189 authors, 9079 institutes, and 6933 venues left.}
provided by the authors of Heterogeneous Graph Transformer (HGT)~\citep{hu2020heterogeneous} in order to have a fair comparison with HGT.
Field nodes are further divided into 6 levels (L0 to L5) and as a result, the
graph comes with 15 edge types, adding rich edge relation information to the
graph. Each paper is featurized with a language embedding generated by pre-trained XLNet on its title. Following HGT, we evaluate two tasks on paper nodes:
predicting publishing venues and predicting L1 field (multi-label). One
potential issue with OAG dataset is indirect information leakage since target
nodes have edges connecting to ground truth label nodes in OAG graph. To address
this issue, for each task, we remove all edges between paper nodes and
corresponding label nodes we are to predict. For example, if the task is
predicting venues of papers, we remove all edges between paper nodes and venue
nodes.

\end{document}